\documentclass[runningheads]{llncs}
\usepackage{cw}

\begin{acronym}[CNN] 
	\acro{CNN}{Convolutional Neural Network}
	\acrodefplural{CNN}[CNNs]{Convolutional Neural Networks}
	\acro{SNN}{Spiking Neural Network}
	\acrodefplural{SNN}[SNNs]{Spiking Neural Networks}
	\acro{TTI}{Time-to-Impact}
	\acro{EDeNN}{Event Decay Neural Network}
	\acrodefplural{EDeNN}{Event Decay Neural Networks}
	\acro{EDeC}{Event Decay Convolution}
	\acro{GNN}{Graph Neural Network}
	\acrodefplural{GNN}{Graph Neural Networks}
	\acro{AEE}{Average Endpoint Error}
	\acro{ANN}{Artificial Neural Network}
	\acro{LIF}{Leaky Integrate and Fire}
	\acro{SRM}{Spike Response Model}
	\acro{SAE}{Surface of Active Events}
\end{acronym}

\graphicspath{{images/},{diagrams/},{diagrams/build/}}
\begin{document}
\title{EDeNN: Event Decay Neural Networks for low latency vision}

\titlerunning{EDeNN: Event Decay Neural Networks for low latency vision}
\author{Celyn Walters\orcidID{0000-0001-6108-4517} \and
Simon Hadfield\orcidID{0000-0001-8637-5054}}
\authorrunning{C. Walters and S. Hadfield}
\institute{CVSSP, University of Surrey, UK
	\email{celyn.walters@surrey.ac.uk}\\
}

\maketitle
\hyphenpenalty=10000

\begin{abstract}\label{abstract}
Despite the success of neural networks in computer vision tasks, digital `neurons' are a very loose approximation of biological neurons.
Today's learning approaches are designed to function on digital devices with digital data representations such as image frames.
In contrast, biological vision systems are generally much more capable and efficient than state-of-the-art digital computer vision algorithms.
Event cameras are an emerging sensor technology which imitates biological vision with asynchronously firing pixels, eschewing the concept of the image frame.
To leverage modern learning techniques, many event-based algorithms are forced to accumulate events back to image frames, somewhat squandering the advantages of event cameras.\looseness=-1

We follow the opposite paradigm and develop a new type of neural network which operates closer to the original event data stream.
We demonstrate state-of-the-art performance in angular velocity regression and competitive optical flow estimation, while avoiding difficulties related to training \aclp{SNN}.
Furthermore, the processing latency of our proposed approach is less than \(1/10\) any other implementation, while continuous inference increases this improvement by another order of magnitude.
\end{abstract}
\acresetall

\section{Introduction}\label{intro}

\begin{figure*}[t]\centering
	\includegraphics[width=\linewidth]{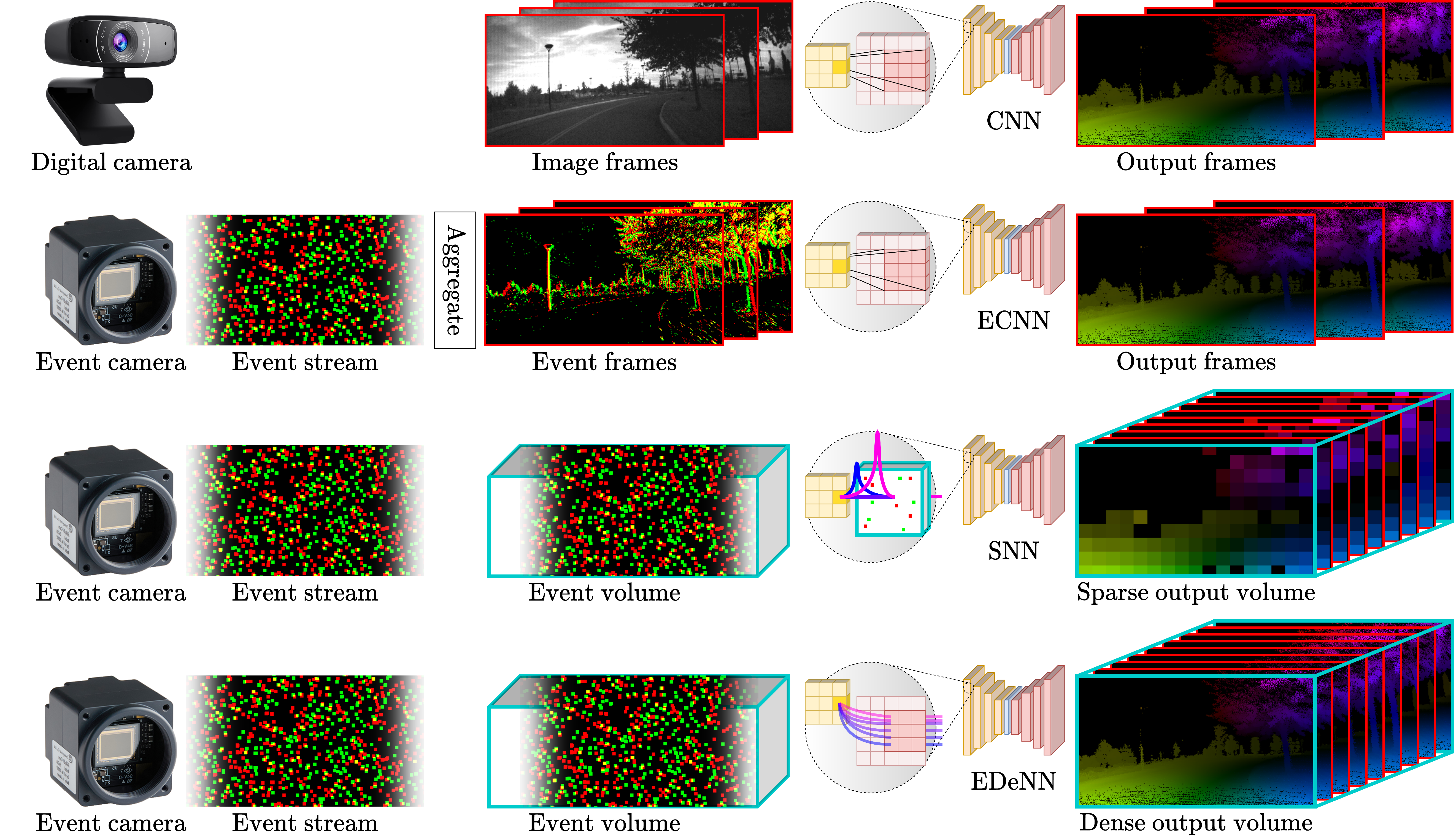}%
	\caption{\label{fig:front}
		A conceptual comparison between various neural networks.
		\textbf{Row 1}: A \ac{CNN} operates on image frames individually to produce output frames.
		\textbf{Row 2}: An event-based \ac{CNN} (denoted as `ECNN') also operates on frames which are aggregated and quantised from the event stream.
		\textbf{Row 3}: An \ac{SNN} operates on events directly and can produce output with high temporal resolution, but cannot usually produce an output at the original spatial resolution.
		\textbf{Row 4}: The proposed \ac{EDeNN} also operates on events directly, but can produce output with both high temporal resolution and full spatial resolution.
	}
\end{figure*}
Event cameras are a neurologically inspired visual sensor with a rising popularity in computer vision research.
They are able to capture data with a much higher temporal resolution than traditional cameras, effectively overcoming motion blur.
Their high dynamic range is also useful for detecting small changes and assists in low-light conditions.
The reduced latency also makes it feasible to react rapidly to external stimulus.
As opposed to traditional frame-based cameras, individual events are streamed asynchronously.
Unfortunately, since the majority of computer vision algorithms assume synchronous pixel measurements, it can be challenging to adapt these algorithms to work with the data stream produced by event cameras.
For this reason, much existing work on event cameras has followed the naïve approach of accumulating the events into image frames at a fixed framerate~\cite{Zhu2018}.
In this paper, we argue that this aggregation neglects the main benefits of this kind of sensor modality.
The sensor's true potential can only be realised by introducing new processing techniques which cater to the characteristics of event streams.
To this end we propose \ac{EDeNN}.
\Cref{fig:front} contrasts our proposed approach against other standard network types.

\begin{figure*}[h]\centering
	\subcaptionbox{\label{fig:neuron:spikes}
		Input spike train which with positive and negative weightings.
	}[\linewidth]{\includegraphics[width=\linewidth]{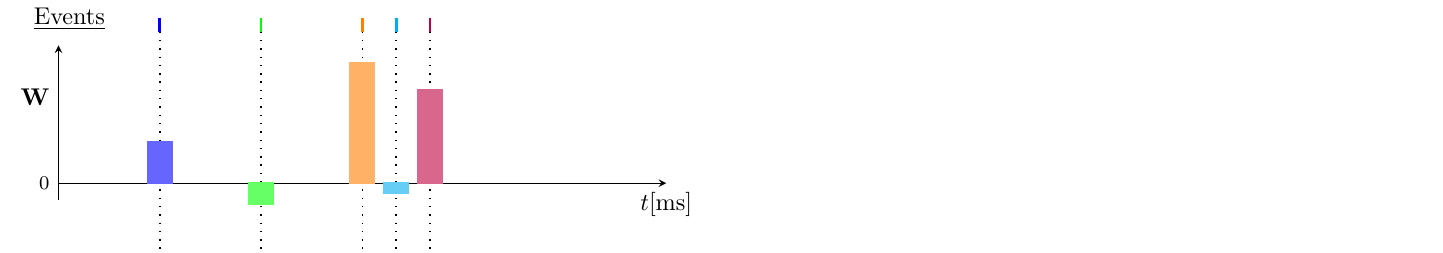}}%

	\subcaptionbox{\label{fig:neuron:snn}
		\ac{SNN} behaviour.
		\textbf{Left}: Black line shows the neuron's membrane potential, which has a lead-in time for each input spike.
		Output spikes are followed by a refractory response.
		\textbf{Right}: Red line shows an output spike when the membrane potential exceeds the neuron threshold.
	}[\linewidth]{\includegraphics[width=\linewidth]{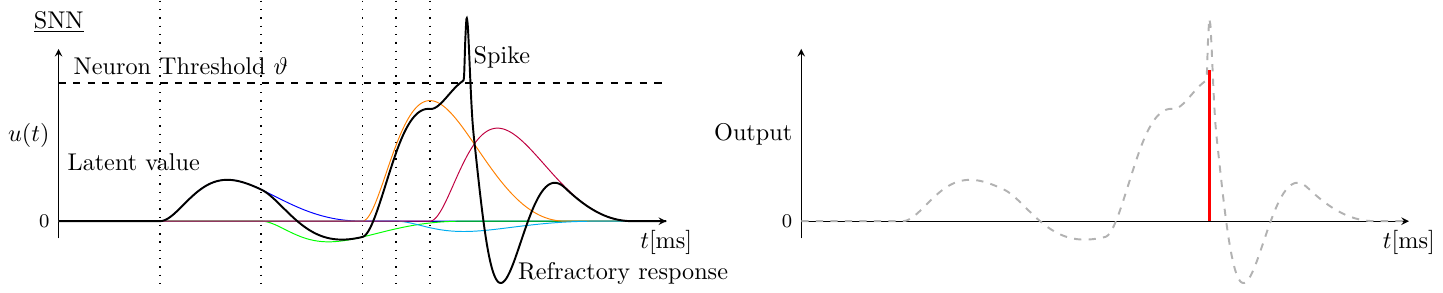}}%

	\subcaptionbox{\label{fig:neuron:ours}
		Behaviour of our proposed \ac{EDeNN} model.
		\textbf{Left}: Black line shows the neuron's latent value
		\textbf{Right}: Red line shows that the continuous latent value comprises the output.
	}[\linewidth]{\includegraphics[width=\linewidth]{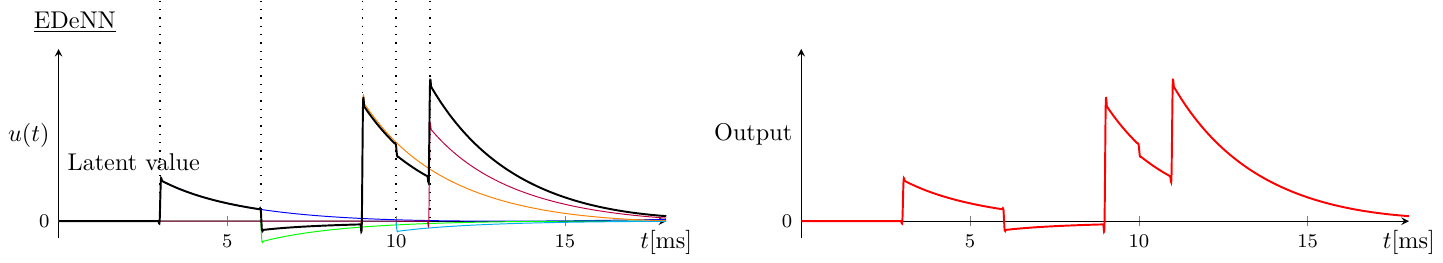}}%
	\caption{\label{fig:neuron}
		Dynamic responses from a single neuron given a weighted 1-dimensional spike train.
	}
\end{figure*}

Despite the fact that \acp{CNN} are conceptually based on neurological structures, both the sensory input and the perception are catered towards synchronous parallel hardware (cameras) and common data storage formats (images).
With a view to treating event streams neuromorphically, our proposed \ac{EDeNN} approach also takes some inspiration from \acp{SNN}.
\Acp{SNN} are designed to more closely approximate biological neuron activity.
In contrast to a \ac{CNN}, each neuron in an \ac{SNN} fires asynchronously and has an impulse response which only propagates discrete neural `spikes' to subsequent neurons.
The series of spikes reaching a given neuron is known as a spike train.
Each spike in the train contributes to the neuron's membrane potential which is a continuously decaying state value.
Once a neuron's membrane potential overcomes a certain threshold, that neuron transmits a spike to neurons in the next layer, followed by a short refractory period where it is less affected by additional incoming spikes.
\Cref{fig:neuron:snn} shows how individual events with positive and negative weightings affect the membrane potential and lead to an output spike train.
Events generated from event cameras are already in the ideal data representation for \acp{SNN}, and an \ac{SNN} layer produces another spike train as output.

Although an \ac{SNN} may be more suited to the data representation of event cameras, in practice it is far more difficult to train an \ac{SNN} than a traditional \ac{CNN}.
This often limits applications to classification tasks~\cite{Chaney2021}.
A neuron can only transmit up to a single output spike for each incoming spike, usually much fewer.
This means that every subsequent layer reduces the amount of
information flowing through the network.
There are clear similarities between this issue and the vanishing gradient problem in traditional \acp{CNN}.
However, for event-driven learning, it is exacerbated by the discretisation of the spike train and the fact that neuron weightings cannot be updated when no output spikes occur.
This significantly slows down the training speed of \acp{SNN} as large portions of the training data become unable to affect many of the network parameters.
It also leads to wasted network capacity due to the `dead neuron' problem:
If a neuron's weightings accidentally reach values where they do not cause output spikes for any of the training samples, that neuron will never have its weights updated and will remain unresponsive forever.
To mitigate this, the majority of \ac{SNN} implementations seek to regress a small number of scalar values from full input images, such as angular velocity regression~\cite{Gehrig2020} and action categorisation~\cite{Shrestha2018}.
By reducing the spatial dimensions in this way, the number of spikes in the output spike train can be maintained by pooling from neighbouring pixels.
\Acp{SNN} are historically poorly suited to learning dense estimation tasks such as optical flow estimation or semantic segmentation.

To fully leverage the advantages of event cameras, our \ac{EDeNN} approach is designed to take the best from both \acp{CNN} and \acp{SNN} in terms of accuracy and latency.
In summary, we make the following contributions.
\begin{enumerate}[label=\textbf{\arabic*)}]
	\item We define \acp{EDeNN}, a new approach to deep learning with event streams.
	Networks built with this paradigm may utilise events directly without accumulating and quantising into low-frequency `event image' frames.
	\acp{EDeNN} also emphasise the latency of decision making and can be deployed in an extremely efficient online streaming mode.
	\item To deal with the sparsity inherent in raw event streams, we propose a new formalisation of partial convolutions which takes learned biases into account.
	This yields an improvement of \SI{23}{\percent} on the original implementation~\cite{Liu2018a}.
	\item To avoid dead neurons and the vanishing gradients/spikes problem, we propose a new activation model which propagates the neuron's latent value directly without the need for discretisation.
	\item We make public a new library, built on the pytorch auto-differentiation engine, which allows researchers to develop and train their own EDeNN solutions to different problems.
\end{enumerate}

\section{Related Work}\label{lit}

Event cameras enable some applications where traditional frame-based cameras would fail, such as those requiring high speeds with rapid reflexes or in low-light conditions.
One area which has seen significant interest is the control of unmanned aerial vehicles, where \citeauthor{Sanket2019} enable dynamic obstacle avoidance on a quadrotor~\cite{Sanket2019}, and \citeauthor{Dietsche2021} track electrical power lines~\cite{Dietsche2021}.

Because event cameras are an emerging technology, there are still comparatively few labelled datasets.
Synthetic or partially synthetic datasets using event camera simulators such as ESIM~\cite{Rebecq2018} are the most common~\cite{Mueggler2017,Walters2021}.
The MVSEC dataset~\cite{Zhu2018a} includes stereo event cameras from multiple vehicles, in indoor and outdoor scenes, recorded in the day and night.
It also provides optical flow, LiDAR, IMU and GPS, meaning it is suitable to be used to evaluate many different tasks, such as visual odometry and depth prediction.
More recently, the DSEC dataset~\cite{Gehrig2021a} additionally incorporates global shutter cameras, disparity maps, and increases the event camera resolutions.
A total of 53 driving sequences are included, suitable for training and evaluating low-light automotive applications.

Before the rise of neural-networks for event data, many approaches were gradient-based.
\Citeauthor{Benosman2014} fit local planes to spatio-temporal event clouds~\cite{Benosman2014}.
An edge moving through space over time manifests as a surface in the 3D volume, commonly referred to as a \ac{SAE}.
The coefficients for the planes fitted to this surface encode the edge motion, enabling the estimation of optical flow.
Methods such as this one require a point neighbourhood which is not too small or large for stable plane-fitting.

Although event cameras are rising in popularity in the field of computer vision, the vast majority of event-based neural networks are architecturally similar to those based on traditional RGB images.
The events are accumulated into frames at a consistent framerate, discarding information.
\Citeauthor{Gehrig2020} argue that the accurate timestamps of events are crucial, and without them the performance degrades significantly~\cite{Gehrig2020}.
Nevertheless, \acp{CNN} act upon frames or volumes, so events must be accumulated in some way before a traditional deep learning tool focused on \acp{CNN} can be applied.
This inherently involves a loss of information as multiple events are combined.
Increasing the temporal resolution of event slices also increases their sparsity and leads to inefficient use of memory, storage, and computation.
As a result, a trade-off is usually made to preserve information and maintain efficiency.
Some \ac{CNN} techniques choose to store the event counts for each pixel location~\cite{Nguyen2019}, while others incorporate the relative event timestamps to try to preserve more temporal information~\cite{Zhu2018}.
Despite many of these papers citing the temporal advantages of this technology, it is rare for authors to measure the processing latency of their learning approaches.
It is therefore not possible to ascertain if this dominates the advantages of the sensor itself, and therefore whether such approaches are useful in practice.
In contrast \acp{EDeNN} focus heavily on latency and efficient inference, and we explicitly contrast this against previous methods in \Cref{results}.

\Acp{SNN} also do not require a lossy information aggregation step, rather they operate directly on the event stream with each layer producing a series of discrete and asynchronous output events.
There are also various approaches to model neuron activity with differing levels of approximation.
The Hodgkin-Huxley neuron~\cite{Hodgkin1952} uses four differential equations to compute the membrane potential, and is considered too computationally complex to use for the large number of neurons comprising an \ac{SNN}.
The \ac{LIF}~\cite{gerstner_kistler_2002} model is commonly used with \acp{SNN} thanks to its much simpler formulation, although it cannot capture all the dynamical behaviours of real neurons.
The \ac{SRM}~\cite{Gerstner1995} model is similar to \ac{LIF}, but additionally includes refractory responses to output spikes (see \cref{fig:neuron:snn}).
One of the main challenges with implementing \acp{SNN} is enabling backpropagation through differentiable functions.
There have been many individual approaches, each formalising the problem differently.
Some overcome the challenge by first training an \ac{ANN} and subsequently converting it to an \ac{SNN}~\cite{Hunsberger2015,OConnor2013,Liu2017}.
This usually negatively impacts the accuracy, although measures are often introduced in an attempt to overcome this.
Many others only backpropagate the membrane potential at certain times, ignoring the temporal dependency between spikes.
\Citeauthor{Shrestha2018} released SLAYER, a CUDA-accelerated software framework to train \acp{SNN} by representing the derivative of spike functions (using the \ac{SRM} model) by a probability density function~\cite{Shrestha2018}.
Unfortunately, the fact that every input spike produces one or fewer output spikes leads to the vanishing gradient/spike and dead neuron problems.
This usually restricts the application of \acp{SNN} to scalar classification and regression problems.
In contrast, an \acp{EDeNN} avoids the spike discretization step, mitigating vanishing gradients and completely eliminating the dead neuron problem.

Outside the field of event-camera processing, there have been some other attempts to apply traditional \ac{CNN} deep learning tools to non-image data.
Notably, PointNet~\cite{Qi2017} and its successors~\cite{Qi2017a,Wu2019,Shi2019} attempt to process point cloud data which, like event data, tends to be extremely sparse.
\acp{GNN} go a step further and eliminate the concept of spatial neighbourhoods in images entirely, replacing it with the concept of connectivity in order to process graph-based data~\cite{Shi2020,Zhang2020}.
Unfortunately, because these techniques were not developed with event cameras in mind, they make no concessions to the efficiency nor latency of their inference.
They generally must have access to the entire point cloud or graph at once before a prediction can be made.
In contrast, our proposed \acp{EDeNN} are capable of extremely efficient low-latency streaming inference.

\section{Methodology}\label{method}

Events originating from a neuromorphic sensor such as an event camera are streamed continuously using a sparse index style notation.
This is generally formatted as \(\mathcal{E} = {[\bm{x}, p, t]}\), with each event comprising image coordinates \(\bm{x} = [x, y]\), polarity \(p\) and timestamp \(t\).
These sparse indices can be used to recreate an event volume \(I \in \mathbb{R}^{W \times H \times C \times T}\) with similar spatial and temporal properties to those common in temporal \acp{CNN}.
\begin{equation}
	I(\bm{x}, p, t) =
	\begin{cases}
		1, & \textnormal{if } [\bm{x}, p, t] \in \mathcal{E}\\
		0, & \textnormal{otherwise}
		.
	\end{cases}
\end{equation}

The building block of an \ac{EDeNN} is the \ac{EDeC} layer.
Inspired by \acp{SNN}, the \ac{EDeC} layer also uses a decaying latent value to find temporal associations between sparsely distributed events.
However, inspired by \acp{CNN} this is combined with a spatial convolution operation to recognise structural scene elements.
Note that unlike an \ac{SNN}, the \ac{EDeC} layer does not discretise its output into a spike train.
The latent value itself is propagated to the next layer, avoiding a decrease in information density leading to vanishing gradients/spikes and dead neurons.

Each \ac{EDeC} layer is comprised of a number of \ac{EDeC} neurons, which independently produce a single channel of the layer's output volume.
In the most general sense, a single \ac{EDeC} neuron performs a 3D spatio-temporal convolution of its input \(I\) and a learned spatio-temporal kernel \(\hat{K} \in \mathbb{R}^{\hat{W} \times \hat{H} \times \hat{T}}\) (note that to help distinguish 2D and 3D convolutions, we indicate the domain of the convolutions using `\(:\)')
\begin{equation}\label{eq:st_conv}
	E(:, \bar{c}, :) = \displaystyle \sum_{c \in C} \hat{K}^{\bar{c}}_c * I(:, c, :)
	.
\end{equation}
However, in an \ac{EDeC} neuron the kernel \(\hat{K}\) is constrained to model only a certain class of functions.
Specifically, each \ac{EDeC} neuron is designed to exhibit the Markov property.
This means that for each new slice in the input volume \(I\), we only need the result for the previous slice and our new input slice.
This makes it possible to perform streaming inference, where every incoming event slice is immediately processed and a prediction computed, without the need to reprocess the event volume.

In this paper, we enforce this by parameterising each \ac{EDeC} neuron as a set of \(K \in \mathbb{R}^{\hat{W} \times \hat{H}}\) spatial parameters plus one temporal decay parameter \(\gamma \in [-1..1]\).
As with \acp{SNN}, the decay rate is a parametric mathematical function.
However, unlike \acp{SNN} we make the parameters of these decay functions learnable, and varying across neurons.
We thus define the spatio-temporal kernel \(\hat{K}\) as
\begin{equation}\label{eq:st_kernel}
	\hat{K}_c(\bm{x}, t) = K_c(\bm{x}) \gamma^{\hat{T} - t}
	.
\end{equation}

\subsection{Filter separability for streaming inference}

Thanks to this particular parameterisation, an \ac{EDeC} neuron's spatio-temporal convolution filter \(\hat{K}\) is linearly separable into two components which are highly effective for event based learning.
These roughly translate to a spatial \ac{CNN} element and a temporal \ac{SNN} component.
To see this, we first note that we can combine equations \cref{eq:st_conv} and \cref{eq:st_kernel} to define the output of an \ac{EDeC} neuron at a particular time, as a weighted summation over previous timesteps
\begin{align}\label{eq:edec_summ}
	E(:, \bar{c}, t) = \sum_{c \in C} \sum_{\tau = t}^{t + \hat{T}} K^{\bar{c}}_c * I(:, c, \tau) \gamma^{\hat{T} - \tau}
	.
\end{align}
Next we move the final item \(\tau = t + \hat{T}\) outside of the summation along with a power of gamma
\begin{multline}
	E(:, \bar{c}, t) = \\
	\sum_{c \in C}\left[K^{\bar{c}}_c * I(:, c, t) + \gamma \sum_{\tau = t}^{t + \hat{T} - 1} K^{\bar{c}}_c * I(:, c, \tau) \gamma^{\hat{T} - \tau - 1}\right]
	.
\end{multline}
Finally, we note that the right hand side of the equation is equivalent to \cref{eq:edec_summ} at \(t - 1\).
This enables us to produce a definition which is recursive in time
\begin{align}
	E(:, \bar{c}, t) = \sum_{c \in C}\left[K^{\bar{c}}_c * I(:, c, t) + \gamma E(:, c, t - 1)\right]
	.
\end{align}

\begin{figure}[h]\centering
	\includegraphics[width=0.7\linewidth]{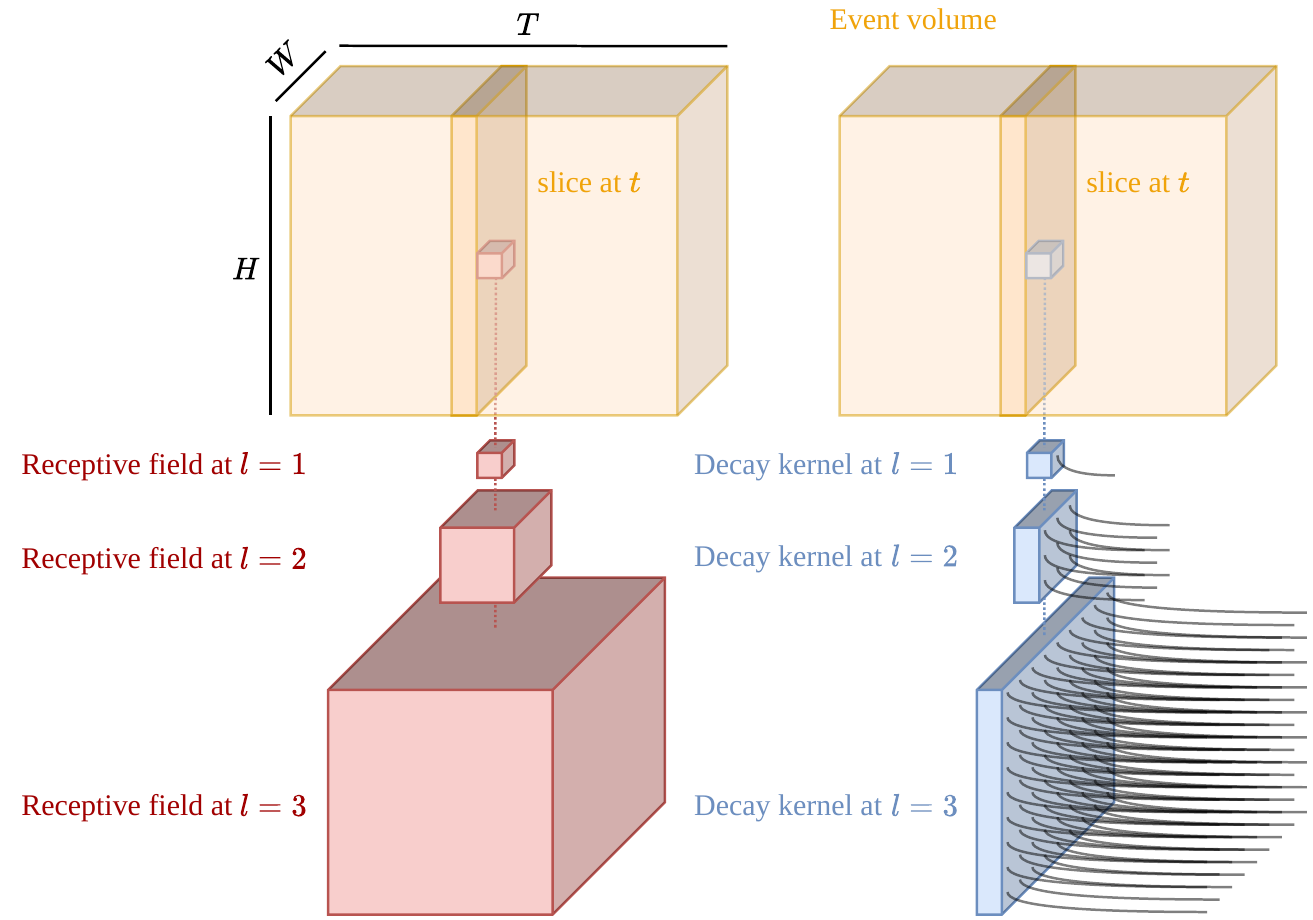}%
	\caption{\label{fig:kernels}
		Receptive fields for traditional 3D convolutional kernel (left) vs.~our proposed learnable Markovian decay kernel (right).
		Temporal relationships can still be learned without the computational and memory constraints of a large volume.
	}
\end{figure}

We can generalize this beyond the first layer of the network.
Thanks to this separability, the forward pass of an \ac{EDeC} neuron at any layer \(l\) and time slice \(t\) can be defined as the sum of two parts.
First, the convolved output of the slice at the same time \(t\) in the previous layer \(l - 1\), and secondly the decayed output of the previous time slice \(t - 1\) at the current layer
\begin{align}\label{eq:forward}
	E^l(:, \bar{c}, t) = \sum_{c \in C}\left[K^{\bar{c}}_c * E^{l - 1}(:, c, t) + \gamma E^{l}(:, c, t - 1)\right]
	.
\end{align}
There are two main advantages of this separable formulation, which drastically improve the computational efficiency of an \ac{EDeNN}.
These are shown in \Cref{fig:kernels}.
Firstly, both components are simple to compute.
The 2D convolution operation has a number of parameters and complexity that scales with the square of the kernel size rather than its cube.
Meanwhile, the second term can be pre-computed before the next time slice arrives.
This helps maintain a low processing latency.
However, the second and more fundamental implication of our separable formulation is the nature of the receptive field at deeper layers.
The Markovian property of the temporal operation ensures that the output at every layer of the network will only depend on, at most, a single previous timestep.
In contrast, for a standard spatio-temporal \ac{CNN} convolution, the receptive field grows for each subsequent layer.
This is problematic, as temporal expansion of the receptive field implies that the \ac{CNN} must wait for future slices to be observed before it is able to make predictions about past slices.
This greatly increases the latency of the prediction, and precludes true streaming inference with \acp{CNN}.

\subsection{Weighted Partial Convolutions}

Although our formulation supports efficient temporal propagation of events, the raw event tensor tends to be very sparse.
As a result, the empty regions dominate the values produced by normal convolutions.
More importantly these empty regions dominate the model's weight updates.
\textit{Partial} convolutions are more suited to these sparse tensors.

Partial convolutions have previously been presented for applications to image inpainting~\cite{Liu2018a} and for zero-padding~\cite{Liu2018b}.
On the first layer of a network, a binary mask \(\mathbf{M}\) is provided alongside the input tensor.
This mask normally represents holes or zero-padding regions.
The convolution operation is subsequently disabled for masked regions, and the unmasked pixels are convolved as if the holes were not present.
This is shown to be beneficial in terms of computational efficiency and training stability.
In the original formalisation by \citeauthor{Liu2018a}, the elements of \(E^{l}\) are masked during convolution, and the resulting elements of \(E^{l+1}\) are multiplied by a scaling factor \(\alpha\) based on the number of surrounding masked cells.
This removes the effect of holes on the unmasked regions.
We extend this idea to \ac{EDeC} neurons as
\begin{align}\label{eq:partial}
	E^{l}(:, \bar{c}, t) = &
	\alpha^{l}(:, t) \sum_{c \in C}
	& \left[K^{\bar{c}}_c * E^{l\shortminus 1}\left(:, c, t\right) \odot \mathbf{M}^{l\shortminus 1}\left(:, t\right) \right.\nonumber\\
	& & \left. + \gamma E^{l}(:, c, t \shortminus 1) \odot \mathbf{M}^l(:, t \shortminus 1) \right]
\end{align}
where \(\odot\) represents the Hadamard product.
The scaling factor \(\alpha\) is defined as
\begin{multline}
	\alpha^{l}(\bm{x}, t) =
	\frac{2 |\Omega| }
	{\displaystyle \sum_{\delta \in \Omega} \left[
	\mathbf{M}^{l \shortminus 1}(\bm{x} \!+\! \delta \bm{x}, t)
	+ \mathbf{M}^l(\bm{x} \!+\! \delta \bm{x}, t \shortminus 1)\right]}
	,\\
\end{multline}
where \(\Omega\) is the domain of 2D offsets within the kernel \(K\).
In the case of a division by zero (\ie~the masks are empty) the value of \(\alpha\) is set to zero.

At the first layer of the \ac{EDeNN}, the mask is determined as the areas where events of either polarity occur in the input
\begin{equation}
	\mathbf{M}^{0}(\bm{x}, t) = I(\bm{x}, 1, t) + I(\bm{x}, \shortminus 1, t)
	.
\end{equation}
For subsequent layers of the \ac{EDeNN}, we define the mask as
\begin{equation}
	\mathbf{M}^{l}(\bm{x}, t) =
	\begin{cases}
		1, & \textnormal{if } \alpha^{l}(\bm{x}, t) > 0\\
		0, & \textnormal{otherwise}
		.
	\end{cases}
\end{equation}

This approach closely mirrors that of the original partial convolution definition for \acp{CNN} in that it successfully ignores the presence of the masked cells by `averaging out' their result.
To be explicit, the value of \(E\) is invariant to the unmasking of spatial element \(\bar{\mathbf{M}}^{l \shortminus 1}\), under the following condition:
\begin{equation}
	\sum_{c \in C} K^{\bar{c}}_c * E^{l \shortminus 1}(:, c, t) \odot \bar{\mathbf{M}}^{l \shortminus 1}
	= \frac{E^{l}(:,{\bar{c}}, t)}{2 |\Omega| }
	.
\end{equation}
Similarly, the value of \(E\) remains unchanged when unmasking temporal element \(\bar{\mathbf{M}}^{l}\) under the following condition:
\begin{equation}
	\sum_{c \in C}\gamma E^{l}(:, c, t \shortminus 1) \odot \bar{\mathbf{M}}^l
	= \frac{E^{l}(:, \bar{c}, t)}{2 |\Omega| }
	.
\end{equation}
Thus we can imagine that the masking operation does not affect the output value, only if we assume that the \textit{impact} of the unobserved elements on the output would be equal to the average impact of the observed elements.

Unfortunately, it is apparent that this definition conflates the effect of the masked inputs and their corresponding kernel values.
If an `important' (\ie~high weight) element of the kernel is masked out, this causes the same change to \(\alpha\) as if an `unimportant' (\ie~low weight) kernel element is masked out.
In fact, the conditions above will only hold true when assuming that the value of the ignored kernel elements are exactly counteracted by the values of the missing input elements.
In the extreme case of a kernel element with weight \num{0} being masked, the conditions above imply that we can only maintain invariance to the masking operating if the masked input item had an infinite value.

Inspired by this observation, we deviate from the original partial convolution formulation and specify an updated scaling factor \(\hat{\alpha}\) which accounts for the learned kernel weights in the masked region
\begin{align}
	\hat{\alpha}^{l}(\bm{x}, \bar{c}, t) & = \frac{\gamma |\Omega| + \displaystyle \sum_{c \in C} \displaystyle \sum_{\delta \in \Omega} K^{\bar{c}}_c(\delta \bm{x})}{\left[a + \gamma b\right]}
	,
\end{align}
where \(a\) is the total unmasked spatial kernel weight
\begin{align}
	a & =
	\displaystyle \sum_{c \in C}
	\displaystyle \sum_{\delta \in \Omega}
	K^{\bar{c}}_c(\delta \bm{x})\mathbf{M}^{l \shortminus 1}(\bm{x} + \delta \bm{x}, t)
	,
\end{align}
and \(b\) is the total unmasked temporal kernel weight
\begin{align}
	b & =
	\displaystyle \sum_{c \in C}
	\displaystyle \sum_{\delta \in \Omega}
	\mathbf{M}^l(\bm{x} + \delta \bm{x}, t \shortminus 1)
	.
\end{align}
Intuitively, the new scaling factor measures the maximum importance of all input elements (measured by their learned kernel weightings) divided by the total importance of the unmasked input elements.
As with the original scaling factor, we note that when all inputs are observed \(\hat{\alpha}\) is simply 1 and when all inputs are unobserved we set \(\hat{\alpha}\) to 0.
However, we can see that under the proposed scheme, masking or unmasking an important (high weight) element would be expected to cause a larger change in the scaling factor.
Meanwhile, (un)masking an unimportant (0 weight) item will no longer cause any change in the output, regardless of the value of corresponding input element.

We believe that these properties make our \ac{EDeC} neuron a better choice for building layers which are
invariant to input masking.
Therefore we expect learned \acp{EDeNN} to generalise more successfully to variations in the masking of the input event volume (\ie~variations in the arrangements of sparse events).

\section{Results}\label{results}

To show the generalisability of our \ac{EDeNN} approach to event streams, we evaluate the latency as well as the accuracy in two different scenarios:
\begin{enumerate}
	\item We estimate angular velocity for a moving camera, comparing directly with both an \ac{SNN} and traditional neural networks,
	This is a traditionally good problem for \acp{SNN} because it is high frequency and the spatial aggregation helps prevent dead neurons.
	We also perform an ablation study for different kernel operators.
	\item We perform dense estimation of optical flow, which is particularly challenging for \acp{SNN}.
	We compare against the current state of the art traditional \acp{CNN} for event-based optical flow estimation.
\end{enumerate}

\subsection{Scalar regression with EDeNNs}\label{results:ang}

\begin{table*}[h]
	\centering
	\footnotesize
	\begin{tabular}{lcccccc}
		\toprule
		Layer type & \ac{EDeC} 1 & \ac{EDeC} 2 & \ac{EDeC} 3 & \ac{EDeC} 4 & \ac{EDeC} 5 & Fully connected\\
		\midrule
		Kernel size & \(3 \times 3\) & \(3 \times 3 \) & \(3 \times 3 \) & \(3 \times 3\) & \(3 \times 3\) & --\\
		Channels & 16 & 32 & 64 & 128 & 256 & --\\
		Stride & 2 & 2 & 2 & 2 & 1 & --\\
		Parameters & 304 & 4.6k & 18.5k & 73.9k & 295k & 768\\
		\bottomrule
	\end{tabular}
	\caption{\label{tab:architecture}
		\ac{EDeNN} model architecture for the angular velocity regression task.
		Total trainable parameters \(= 394k\).
	}
\end{table*}

\Citeauthor{Gehrig2020} present a dataset and angular velocity regression approach~\cite{Gehrig2020} using the SLAYER \ac{SNN} framework~\cite{Shrestha2018}.
The dataset simulates an event camera being shaken at different rates, with challenging saccadic motion.
Each sample is \SI{100}{\ms} in length at a spatial resolution of \(240 \times 180\).
\Citeauthor{Gehrig2020}'s approach was shown to be competitive with \ac{CNN} baselines with the same number of layers.
Notably, ResNet\=/50, a much deeper network gets the highest performance.
Although they note that deeper \acp{SNN} \textit{may} have better performance, it is also likely to be more challenging to train, due to the issues with vanishing gradients.
In our evaluation we sought to keep as many aspects of the network architecture the same in our \ac{EDeNN}, to allow for a more direct comparison.
Our model has an equivalent configuration of layers, as shown in \cref{tab:architecture}.
It was trained for 500 epochs with a batch size of 4, and we match the settling time allowance protocol of \cite{Gehrig2020} by evaluating the loss from \SI{50}{\ms}~onwards in each window.
\begin{table}[h]
	\centering
	\footnotesize
	\begin{tabular}{
		p{3.45cm}
		c
		S[table-auto-round,table-format=1.2] 
		S[table-auto-round,table-format=1.2] 
		S[table-auto-round,table-format=1.2] 
		c
	}
		\toprule
		Approach & Data & \multicolumn{1}{c}{Relative} & \multicolumn{1}{c}{RMSE} & \multicolumn{1}{c}{Step}\\
		           &  & \multicolumn{1}{c}{error} & \multicolumn{1}{c}{} & \multicolumn{1}{c}{time}\\
		\midrule
		mean& \multicolumn{1}{c}{--} & 1.00 & 226.9 & \multicolumn{1}{c}{--}\\
		ANN\=/6 & \multicolumn{1}{c}{V} & 0.22 & 59.0 & \multicolumn{1}{c}{--}\\
		ResNet\=/50~\cite{He2016} & \multicolumn{1}{c}{A} & 0.22 & 66.8 & \multicolumn{1}{c}{--}\\
		ResNet\=/50~\cite{He2016} & \multicolumn{1}{c}{V} & 0.15 & 36.8 & \multicolumn{1}{c}{--}\\
		SNN\=/6~\cite{Gehrig2020} & \multicolumn{1}{c}{E} & 0.25916576385498047 & 66.32438118067957 & 0.15047207474708557\\
		\midrule
		EDeNN & \multicolumn{1}{c}{E} & \bfseries 0.12 & \bfseries 27.99 & \itshape 0.08956715846673036\\
		\bottomrule
	\end{tabular}
	\caption{\label{tab:ang}
		Comparison of angular velocity regression approaches.
		The data structures (E, A, V) are event-based~\cite{Gehrig2020}, accumulation-based~\cite{Maqueda2018} and voxel-based~\cite{Gehrig2019}, respectively.
		RMSE is in (\degree/\si\second), step times are in (\si\second).\vspace{-1cm}
	}
\end{table}

As shown in \cref{tab:ang}, our approach is clearly more accurate than all other approaches.
We have less than half of both the relative error and RMSE compared to the \ac{SNN}.
Indeed we are even able to clearly outperform much deeper traditional \acp{CNN} such as ResNet\=/50.
In addition to our increased accuracy, we also report a much lower average processing time over the test set than the state-of-the-art \ac{SNN}.
We should note that in order to provide the fairest possible comparison, this time was computed for a cold start over the entire \SI{100}{\ms} sample.
In reality, an \ac{EDeNN}, like an \ac{SNN}, is able to stream new events at inference time.
This means that once the network is primed, subsequent estimations could be performed at a fraction of this cost.
Furthermore, this is all possible without modifying or pre-processing the event data through accumulation (A) or voxelisation (V).

\footnotetext{
	Values taken from~\cite{Gehrig2020}.
	\label{fn:snnpaper}
}

\subsection{Dense estimation with EDeNNs}\label{results:flow}

We next demonstrate the capabilities of \acp{EDeNN} for dense estimation tasks.
These are commonplace with traditional \acp{CNN} but are particularly challenging for \acp{SNN} and implementations of this are extremely scarce.

For this task our \ac{EDeNN} architecture is inspired by EV\=/FlowNet~\cite{Zhu2018}.
The overall architecture comprises four encoder and four decoder layers, and is arranged as a U\=/Net with concatenated residual connections.
Intermediate optical flow estimates of differing resolutions are produced on the output of each decoder layer.
Each of these intermediate flow estimations contributes to the overall loss function with equal weighting.

\nocite{Scheerlinck2020}
\begin{figure}[t]\centering
	\includegraphics[width=0.9\linewidth]{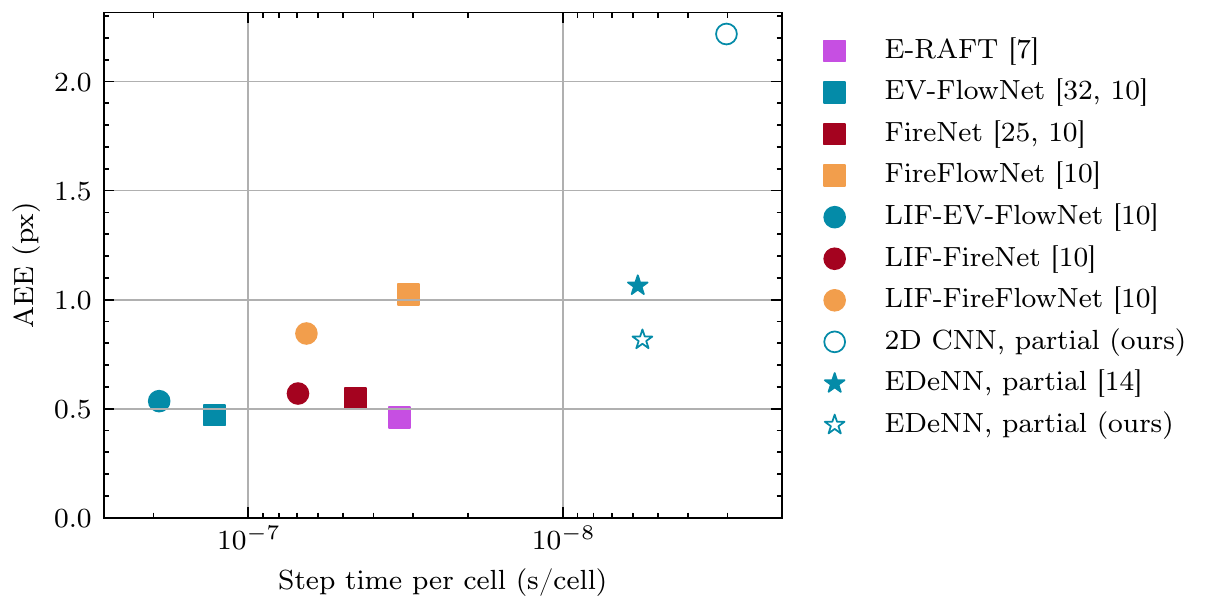}%
	\caption{\label{fig:plot}
		Evaluation of approaches on the `outdoor\_day1' sequence from the MVSEC~\cite{Zhu2018a} dataset.
		\acs{AEE} is the \acl{AEE} in pixels.
		Step time per cell is the average processing time divided by the number of elements in the input data.
		Square markers are \acp{CNN}, round are \acp{SNN}, and stars are \acp{EDeNN}.
	}
\end{figure}

We use a window size of \SI{48}{\milli\second} with bins of \SI{2}{\milli\second} to correlate with the ground truth frame rate of \SI{20}{Hz}, and evaluate the flow predictions of the final layer.
By nature, event cameras do not produce events where there are no lighting or texture changes.
The result of this is that it is almost impossible to estimate optical flow for subjects such as uniformly-coloured walls.
Following the evaluation protocol of previous approaches, performance is computed after masking out pixels where there is no ground-truth flow or where there are no input events.
We use a pixelwise L1 loss to train our \ac{EDeNN}.

\begin{figure}[t]\centering
	\includegraphics[width=0.33\linewidth]{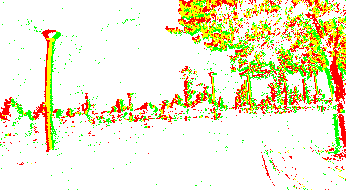}%
	\hfill
	\includegraphics[width=0.33\linewidth]{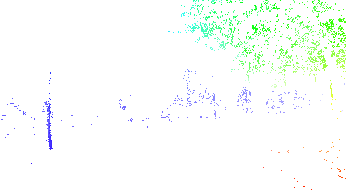}%
	\hfill
	\includegraphics[width=0.33\linewidth]{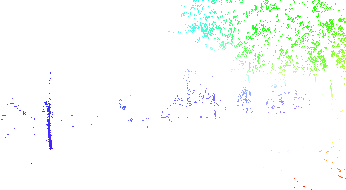}%

	\includegraphics[width=0.33\linewidth]{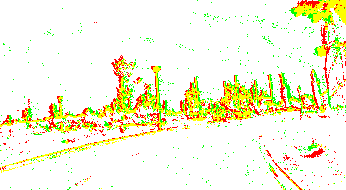}%
	\hfill
	\includegraphics[width=0.33\linewidth]{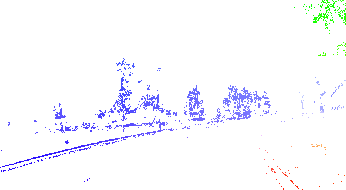}%
	\hfill
	\includegraphics[width=0.33\linewidth]{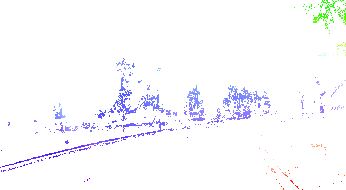}%

	\includegraphics[width=0.33\linewidth]{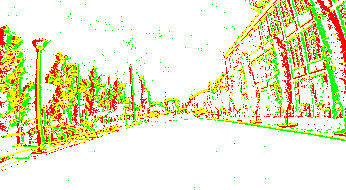}%
	\hfill
	\includegraphics[width=0.33\linewidth]{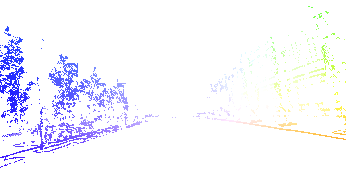}%
	\hfill
	\includegraphics[width=0.33\linewidth]{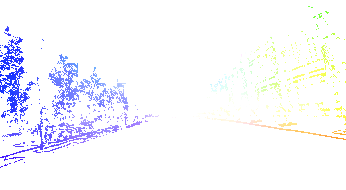}%

	\includegraphics[width=0.33\linewidth]{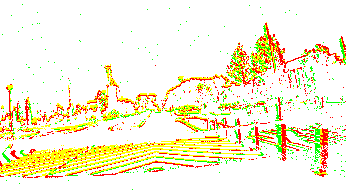}%
	\hfill
	\includegraphics[width=0.33\linewidth]{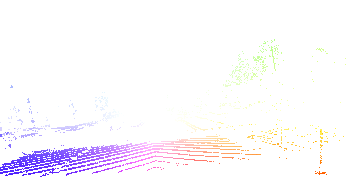}%
	\hfill
	\includegraphics[width=0.33\linewidth]{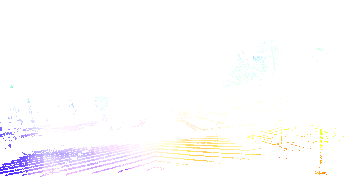}%

	\subcaptionbox{\label{fig:flow:input}
		Events
	}[0.33\linewidth]{\includegraphics[width=\linewidth]{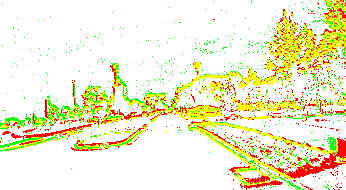}}%
	\hfill
	\subcaptionbox{\label{fig:flow:gtmask}
		Masked ground truth
	}[0.33\linewidth]{\includegraphics[width=\linewidth]{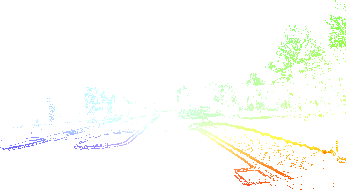}}%
	\hfill
	\subcaptionbox{\label{fig:flow:edenn}
		\ac{EDeNN} (ours)
	}[0.33\linewidth]{\includegraphics[width=\linewidth]{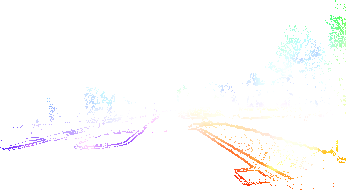}}%
	\caption{\label{fig:flow}
		Results from the optical flow estimation task.
	}
\end{figure}
We estimate optical flow on the MVSEC dataset~\cite{Zhu2018a} and compare against many recent approaches.
The results are shown quantitatively in \cref{fig:plot}.
The step times are measured on identical hardware and averaged over the sequence.
We note that the \ac{SNN} approaches~\cite{Hagenaars2021} tend to have a slightly higher \ac{AEE} and a slightly worse step time per cell than their equivalent \ac{CNN} counterparts.
This is likely due to the vanishing gradient/spike issues which \acp{SNN} experience for dense prediction tasks and to an underexploitation of parallel computing architectures compared to the other techniques.
Overall the top previous state-of-the-art approach appears to be E\=/RAFT~\cite{Gehrig2021b} (evaluated on the original ground truth framerate and resolution) which achieves accuracy comparable to the EV\=/FlowNet baseline, but in a step time which is \num{6} times lower.
In contrast, our \ac{EDeNN} approach, which has a similar layer structure to the original EV\=/FlowNet, is able to further decrease step time by a factor of \num{6} over E\=/RAFT.
Our proposed \ac{EDeC} neuron provides an error reduction of \SI{63}{\percent} over using standard 2D convolutions (`2D \acs{CNN}'),
while our proposed reformulation of partial convolutions yields an error reduction of \SI{23}{\percent} over the original formulation (`\acs{EDeNN} partial~\cite{Liu2018a}').

Qualitative results for the \ac{EDeNN} are shown in \cref{fig:flow}.
We can see that generally speaking the orientation of the estimated flows (shown as hue) match those of the ground truth.
The hue difference on the fourth row of \cref{fig:flow} is located on the road, an image region which does not usually contain events in this dataset.
It is also very interesting to note that our \ac{EDeNN} flow network performs very well at complex structures with fine details such as tree foliage, which is an area where traditional optical flow techniques struggle.
This suggests there is likely a strong synergy between traditional image based \acp{CNN} and \acp{EDeNN}.

\section{Conclusions}\label{conc}

This work proposes a new kind of neural network called \aclp{EDeNN} to overcome limitations of both \acp{CNN} and \acp{SNN} for event-based data.
Because \acp{EDeNN} operate closer to the original event data stream, event accumulation is not necessary which preserves the high temporal resolution and lack of motion blur.
Also, \acp{EDeNN} avoid the dead neurons and vanishing gradients/spikes problem of \acp{SNN}, assisting training for full-sized image outputs.
This is possible by using a new \acl{EDeC} neuron which propagates continuous decaying latent values.
We also propose a new formalisation of partial convolutions which caters to sparse event data by accounting for learned biases.

We showed that \acp{EDeNN} outperform the \ac{SNN} and \ac{CNN} baselines for angular velocity regression for both accuracy and step time.
We also achieve competitive accuracy on optical flow estimation at full resolution, but with orders of magnitude reduction in processing times.
The Markovian nature of \ac{EDeC} kernels suggests that minimal calculation is required for subsequent events, enabling streaming inference and vastly reducing the practical latency/computational complexity resources on real hardware.

\bibliography{bibliography}
\bibliographystyle{icml2023}
\end{document}